\documentclass[sigconf]{acmart}
\usepackage{blindtext}
\usepackage{graphicx}
\usepackage{bbm}
\usepackage{mathtools}
\usepackage{mathrsfs}
\usepackage{diagbox}
\usepackage{multirow}
\usepackage{algorithm}
\usepackage{algorithmic}
\usepackage{bbding}
\usepackage{epsfig}
\usepackage{cases} 
\usepackage{overpic}
\usepackage{subfigure}
\usepackage{color}
\usepackage{hyperref}
\usepackage{makecell}

\usepackage{tabularx}
\usepackage[export]{adjustbox}

\AtBeginDocument{%
  \providecommand\BibTeX{{%
    \normalfont B\kern-0.5em{\scshape i\kern-0.25em b}\kern-0.8em\TeX}}}

\copyrightyear{2021}
\acmYear{2021}
\setcopyright{acmcopyright}\acmConference[MM '21]{Proceedings of the 29th ACM
International Conference on Multimedia}{October 20--24, 2021}{Virtual Event, China}
\acmBooktitle{Proceedings of the 29th ACM International Conference on Multimedia
(MM '21), October 20--24, 2021, Virtual Event, China}
\acmPrice{15.00}
\acmDOI{10.1145/3474085.3475562}
\acmISBN{978-1-4503-8651-7/21/10}

\begin{document}
\fancyhead{}
\title{Memory-Augmented Deep Unfolding Network \\ for Compressive Sensing}

\author{Jiechong Song}
\affiliation{%
  \institution{Shenzhen Graduate School, Peking University}
    \city{Shenzhen}
 \country{China}
}

\author{Bin Chen}
\affiliation{%
  \institution{Shenzhen Graduate School, Peking University}
    \city{Shenzhen}
 \country{China}
}

\author{Jian Zhang}
\authornotemark[1]
\email{zhangjian.sz@pku.edu.cn}
\affiliation{%
  \institution{Shenzhen Graduate School, Peking University}
  \institution{Peng Cheng Laboratory}
  \city{Shenzhen}
  \country{China}
}





\renewcommand{\shortauthors}{Trovato and Tobin, et al.}

\begin{abstract}
Mapping a truncated optimization method into a deep neural network, deep unfolding network (DUN) has attracted growing attention in compressive sensing (CS) due to its good interpretability and high performance. Each stage in DUNs corresponds to one iteration in optimization. By understanding DUNs from the perspective of the human brain's memory processing, we find there exists two issues in existing DUNs. One is the information between every two adjacent stages, which can be regarded as short-term memory, is usually lost seriously. The other is no explicit mechanism to ensure that the previous stages affect the current stage, which means memory is easily forgotten. To solve these issues, in this paper, a novel DUN with persistent memory for CS is proposed, dubbed \textbf{M}emory-\textbf{A}ugmented \textbf{D}eep \textbf{U}nfolding \textbf{N}etwork (\textbf{MADUN}). We design a memory-augmented proximal mapping module (MAPMM) by combining two types of memory augmentation mechanisms, namely High-throughput Short-term Memory (HSM) and Cross-stage Long-term Memory (CLM). HSM is exploited to allow DUNs to transmit multi-channel short-term memory, which greatly reduces information loss between adjacent stages. CLM is utilized to develop the dependency of deep information across cascading stages, which greatly enhances network representation capability. Extensive CS experiments on natural and MR images show that with the strong ability to maintain and balance information our MADUN outperforms existing state-of-the-art methods by a large margin. The source code is available at https://github.com/jianzhan\\gcs/MADUN/.
\end{abstract}
\thanks{This work was supported in part by National Natural Science Foundation of China (61902009). 

$^{\star}$ Corresponding author.}





\maketitle

\section{Introduction}


Compressive sensing (CS) is a novel methodology of acquisition and reconstruction. Signal is first sampled and compressed with linear random transformations. Then, the original signal can be reconstructed from far fewer measurements than required by the sub-Nyquist sampling rate \cite{sankaranarayanan2012cs}\cite{liutkus2014imaging}. Due to its attractive merits of the simple/fast sampling process and the low demand for data transmission and storage, the CS method has spawned many applications, including but not limited to single-pixel imaging \cite{duarte2008single}\cite{rousset2016adaptive}, accelerated magnetic resonance imaging (MRI) \cite{lustig2007sparse}, wireless remote monitoring \cite{zhang2012compressed}, and snapshot compressive imaging \cite{wu2021spatial}\cite{wu2021ddun}.

Mathematically, a random linear measurements $\mathbf{y}\in\mathbb{R}^M$ can be formulated as $\mathbf{y} = \mathbf{\Phi}\mathbf{x}$, where $\mathbf{x}\in\mathbb{R}^N$ is the original signal and $\mathbf{\Phi}\in\mathbb{R}^{M \times N}$ is the measurement matrix with $M\ll N$. $\frac{M}{N}$ is the CS ratio. Obviously, CS reconstruction is an ill-posed inverse problem. To obtain a reliable reconstruction, the conventional CS methods commonly solve an energy function,
\begin{align} \label{eq: opt}
\begin{split}
\setlength{\abovedisplayskip}{5pt}
\setlength{\belowdisplayskip}{6pt}
\underset{\mathbf{x}}{\arg\min} ~~\frac{1}{2} \left \|\mathbf{\Phi}\mathbf{x} - \mathbf{y} \right \|^2_2+ \lambda \mathcal{F}(\mathbf{x}),
\end{split}
\end{align}
where $\lambda \mathcal{F}(\mathbf{x})$ denotes a prior term with regularization parameter $\lambda$. 
For the traditional CS methods \cite{kim2010compressed}\cite{Li2013AnEA}\cite{zhang2014group}\cite{zhang2014image}\cite{gao2015block}\cite{Metzler2016FromDT}\cite{zhao2018cream}, the prior term can be the sparsifying operator corresponding to some pre-defined transform basis, such as discrete cosine transform (DCT) and wavelet \cite{zhao2014image}\cite{zhao2016video}. They enjoy the advantages of strong convergence and theoretical analysis in most cases, but usually inevitably suffer from high computational complexity and face the trouble of choosing optimal transforms and parameters \cite{zhao2016nonconvex}.  

\begin{figure*}[t]
\centering
\includegraphics[width=1.0\textwidth]{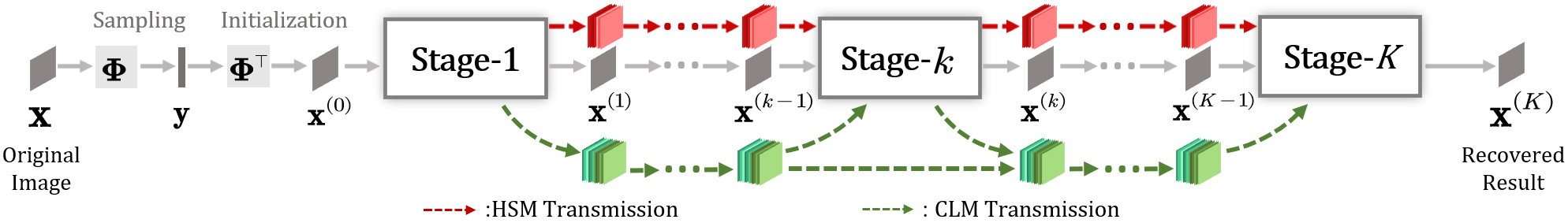} 
\caption{ Architecture of our proposed MADUN, which consists of $K$ stages. $\mathbf{x}$ denotes the full-sampled image for training, $\mathbf{\Phi}$ denotes measurement matrix, $\mathbf{y}$ is the under-sampled data, $\mathbf{x}^{(0)}$ denotes the initialization, $\mathbf{x}^{(K)}$ denotes the recovered result of MADUN and $\mathbf{x}^{(k)}$ stands for the output images of ($k$)-th stage. The dotted red line denotes High-throughput Short-term Memory (HSM) transmission and the dotted green line denotes Cross-stage Long-term Memory (CLM) transmission. 
}
\label{fig:arch}
\vspace{-6pt}
\end{figure*}

Recently, fueled by the powerful learning ability of deep networks, several deep network-based image CS reconstruction algorithms have been proposed \cite{Kulkarni2016ReconNetNR}\cite{sun2020dual}\cite{gilton2019neumann}\cite{chenlearning}, which can be generally grouped into two categories: deep non-unfolding networks (DNUNs) and deep unfolding networks (DUNs). For DNUN, it aims to directly learn the inverse mapping from the CS measurement domain to the original signal domain \cite{mousavi2015deep}\cite{iliadis2018deep}. For DUN, it combines deep network with optimization and trains a truncated unfolding inference through an end-to-end learning process \cite{zhang2018ista}\cite{zhang2020optimization}\cite{you2021ista}\cite{you2021}, which has become the mainstream for CS.  

DUN is usually composed of a fixed number of stages where each stage is mainly influenced by its direct former one, and the mechanism can be considered as short-term memory, \textit{e.g.}, ISTA-Net \cite{zhang2018ista} and DPDNN \cite{zhang2020deep}. In each stage, the information transmission is usually hampered by the feature transformation with channel number reduction from multiple to one. Besides, the information in the prior stages is easily forgotten and the long-term dependency problem is rarely realized, which causes inadequate recoveries.



To address the above issues, in this paper, we propose a Memory-Augmented Deep Unfolding Network (MADUN), focusing on CS reconstruction. In our MADUN, we design a Memory-Augmented Proximal Mapping Module (MAPMM) which contains two different memory augmentation mechanisms, as shown in Figure~\ref{fig:arch}. To reduce information loss between each two adjacent stages, we design a High-throughput Short-term Memory mechanism (HSM) to adaptively add multi-channel information and well ensure maximum signal flow. Also, a Cross-stage Long-term Memory mechanism (CLM) is exploited to explicitly construct the deep and adjustable long-term information path across stages. Our MADUN is a comprehensive framework, considering the merits of previous optimization-inspired DUNs and meanwhile, improving the network representation ability by augmenting entire information transmission among all stages. It enjoys the advantages of both the satisfaction of interpretability and the ensurance of information abundance.
The major contributions are summarized as follows:  
\begin{itemize}
\item
We propose a novel \textbf{M}emory-\textbf{A}ugmented \textbf{D}eep \textbf{U}nfolding \textbf{N}etwork (\textbf{MADUN}) with persistent memory for CS reconstruction, which is able to adaptively capture the adequate features and recover more details and textures. 
\item
We design a \textbf{M}emory-\textbf{A}ugmented \textbf{P}roximal \textbf{M}apping \textbf{M}odule (\textbf{MAPMM}) to enhance information transmission, which contains two different memory augmentation mechanisms. 
\item
We introduce a \textbf{H}igh-throughput \textbf{S}hort-term \textbf{M}emory mechanism (\textbf{HSM}) which is exploited to allow MADUN to transmit high-throughput information between adjacent stages.
\item
We also develop a \textbf{C}ross-stage \textbf{L}ong-term \textbf{M}emory mechanism (\textbf{CLM}) to explore the long-term dependency of deep information across all cascading stages.
\item
Extensive experiments show that, with the strong ability of the balance between the short-term and the long-term memories, our MADUN outperforms existing state-of-the-art networks by large margins.
\end{itemize}

\section{related work}
\subsection{Deep Non-Unfolding Network}
Deep non-unfolding network (DNUN) directly learns mapping functions from the CS sampling image $\mathbf{\Phi}\mathbf{x}$ to the full-sampled image $\mathbf{x}$, resulting in an end-to-end network. 
Kulkarni \textit{et al.} \cite{Kulkarni2016ReconNetNR} develop a CNN-based CS algorithm, dubbed ReconNet, which learns to regress an image block from its CS measurement. Sun \textit{et al.} \cite{sun2020dual} design a dual-path network to learn the structure-texture decomposition in a data-driven manner. Residual learning and U-Net structure are adopted in CS-MRI to successfully learn the aliasing artifacts \cite{Hyun_2018}. Some works jointly learn the measurement by a sampling sub-network and the recovery by a reconstruction sub-network from the training data. CSNet \cite{shi2019image} can avoid blocking artifacts by learning an end-to-end mapping between measurements and the whole reconstructed images. Shi \textit{et al.} \cite{shi2019scalable} exploit a convolutional neural network for scalable sampling and quality scalable reconstruction. 

Apparently compared with traditional methods, DNUNs can represent image information flexibly with the fast inferencing efficiency. However, in DNUNs, the learning performance seriously depends on the careful tuning and the sampling matrix is not well-embedded in the reconstruction process, which not only result in the very tricky training schemes but also drag down the network performance due to the difficulty of directly learning the recovery mapping without explicit matrix guidance.

\subsection{Deep Unfolding Network}
Deep unfolding networks (DUNs) have been proposed to solve different image inverse tasks, such as denoising \cite{chen2016trainable}\cite{lefkimmiatis2017non}, deblurring \cite{kruse2017learning}\cite{wang2020stacking}, and demosaicking \cite{kokkinos2018deep}. DUN has friendly interpretability on training data pairs $\{(\mathbf{y}_j, \mathbf{x}_j)\}_{j=1}^{N_a}$, which is usually formulated on CS construction as the bi-level optimization problem:
\begin{numcases}{}
\begin{aligned} 
\label{eq:DUN-1}
&\underset{\Theta}{\min} \sum_{j=1}^{N_a} {\mathcal{L}(\mathbf{\hat{x}}_j, \mathbf{x}_j)}, \\
\label{eq:DUN-2}
&\mathbf{s.t.}\ \mathbf{\hat{x}}_j=\underset{\mathbf{x}}{\arg \min } \frac{1}{2}||\mathbf{\Phi}\mathbf{x}-\mathbf{y}_j||_{2}^{2}+\lambda\mathcal{F}(\mathbf{x}).
\end{aligned} 
\end{numcases}
\begin{figure*}[t]
\centering
\includegraphics[width=1\textwidth]{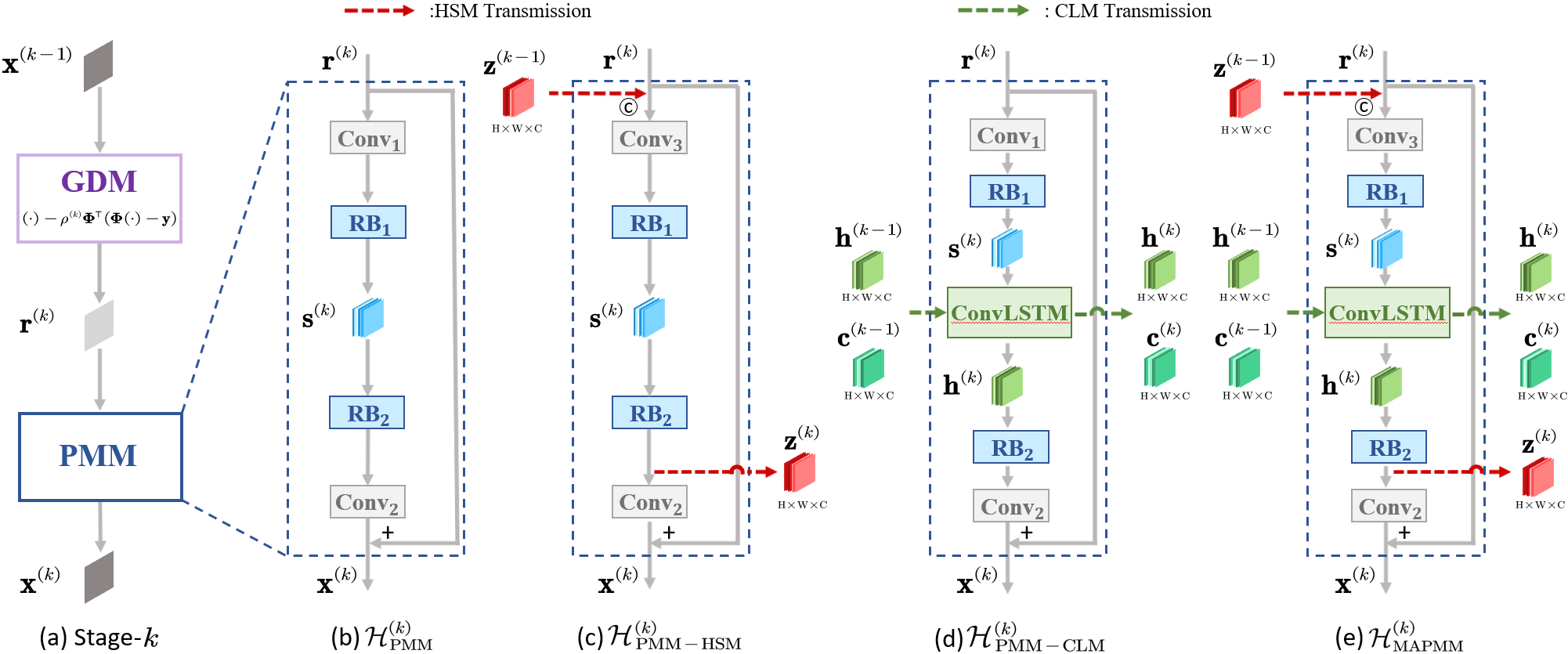} 
\caption{(a) is the illustration of ($k$)-th stage in MADUN, where $\mathbf{r}^{(k)}$ is the outputs of the Gradient Descent Module (GDM) and the inputs of the Proximal Mapping Module (PMM). (b) is the PMM. (c) is the PMM with High-throughput Short-term Memory (HSM). (d) is the PMM with Cross-stage Long-term Memory (CLM). (e) is our proposed Memory-Augmented Proximal Mapping Module (MAPMM) which combines HSM and CLM with PMM. Here, ``+'' is a residual learning strategy, ``$\copyright$'' is the concatenation along channel dimension, $H$ and $W$ denote the height and width of the image and $C$ is the filter number.  
}
\label{fig:stage}
\vspace{-6pt}
\end{figure*}

DUNs on CS and compressive sensing MRI (CS-MRI) usually integrate some effective convolutional neural network (CNN) denoisers into some optimization methods including half quadratic splitting (HQS) \cite{zhang2017learning}\cite{dong2018denoising}\cite{aggarwal2018modl}, alternating minimization (AM) \cite{8067520}\cite{Sun2018CompressedSM}\cite{Zheng2019CascadedDD}, iterative shrinkage-thresholding algorithm (ISTA) \cite{zhang2018ista}\cite{gilton2019neumann}\cite{zhang2020optimization}, approximate message passing (AMP) \cite{zhang2020amp}\cite{zhou2020multi},  alternating direction method of multipliers (ADMM) \cite{yang2018admm} and inertial proximal algorithm for nonconvex optimization (iPiano) \cite{su2020ipiano}. 
Different optimization methods usually lead to different optimization-inspired DUNs. 

Although existing DUNs inherit a good structure from optimization and exhibit friendly interpretability, we find that, from the perspective of neural networks, DUNs' inherent design of taking one-channel images as inputs and outputs for each stage will greatly hinder network information transmission and lose much more image details. Chen \textit{et al.} \cite{chenlearning} propose a contextual memory (CM) module which can augment the sequential links across stages. However, the CM module is still hard to maintain abundant details due to the poor storage capacity of one-channel stage outputs.

\section{Proposed Method}
In this section, we will elaborate on the design of our proposed Memory-Augmented Deep Unfolding Network (MADUN) for CS.

\subsection{Basic Network Architecture}
As we all know, DUN is a kind of CNN method that combines an efficient iterative algorithm. Iterative shrinkage-thresholding algorithm (ISTA) is well suited for solving many large-scale linear inverse problems \cite{zhang2018ista}, \textit{e.g.}, general CS and CS-MRI. Traditional ISTA solves the CS reconstruction problem in Eq.~(\ref{eq: opt}) by iterating between the following update steps: 
\begin{numcases}{}
\label{eq:r}
\mathbf{r}^{(k)}=\mathbf{x}^{(k-1)} - \rho \mathbf{\Phi^{\top}} \left(\mathbf{\Phi} \mathbf{x}^{(k-1)} - \mathbf{{y}}\right), \\
\label{eq:x}
\mathbf{x}^{(k)}=\underset{\mathbf{x}}{\arg \min } \frac{1}{2}||\mathbf{x}-\mathbf{r}^{(k)}||_{2}^{2}+\lambda\mathcal{F}(\mathbf{x}),
\end{numcases}
where $\rho$ is the step size. 
Inspired by ISTA-Net $^+$ \cite{zhang2018ista} which is a DUN unfolded by ISTA, Eq.~(\ref{eq:r}) and Eq.~(\ref{eq:x}) can be expressed as two modules in the ($k$)-th stage $~(0< k \leq K)$ as shown in Figure~\ref{fig:stage} (a): 
\begin{numcases}{}
\label{eq:ista-r}
\mathbf{r}^{(k)}=\mathbf{x}^{(k-1)} - \rho^{(k)} \mathbf{\Phi^{\top}} \left(\mathbf{\Phi} \mathbf{x}^{(k-1)} - \mathbf{{y}}\right), \\
\label{eq:ista-x}
\mathbf{x}^{(k)}=\displaystyle{\mathcal{H}}_{\scriptscriptstyle{\operatorname{PMM}}}^{(k)}(\mathbf{r}^{(k)}).
\end{numcases}
Eq.~(\ref{eq:ista-r}) is called gradient descent module (GDM) where $\rho^{(k)}$ is a learnable parameter, and Eq.~(\ref{eq:ista-x}) is called proximal mapping module (PMM) which is actually a CNN-based denoiser.

To keep a simple structure with the high recovery accuracy, PMM uses basic convolution layers (Conv) and residual blocks (RB) which generate residual outputs by the structure of Conv-ReLU-Conv, and adopts an efficient reconstruction network \cite{zhang2017beyond} as shown in Figure~\ref{fig:stage} (b), which can be divided into four parts:
\begin{itemize}
\item
A convolution layer receives one-channel network inputs $\mathbf{r}^{(k)}$ and generates multi-channel outputs, yielding $\operatorname{Conv_1}(\mathbf{r}^{(k)})$.
\item
A mapping layer extracts deep representation, which consists of two residual blocks ($\operatorname{RB_1}(\cdot)$ and $\operatorname{RB_2}(\cdot)$) in PMM due to optimizing easily \cite{he2016deep}. Here, $\mathbf{s}^{(k)}$ is the intermediate result produced between the two residual blocks.
\item
A convolution layer $\operatorname{Conv_2}(\cdot)$ outputs residual result by the feature conversions from multi-channel to one-channel.
\item
A residual learning strategy produces final module outputs. 
\end{itemize}

Accordingly, PMM can be formulated $\mathbf{x}^{(k)} = \displaystyle{\mathcal{H}}_{\scriptscriptstyle{\operatorname{PMM}}}^{(k)}(\mathbf{r}^{(k)})$ as:
\begin{equation}
\label{eq:PMM}
\mathbf{x}^{(k)} = \mathbf{r}^{(k)} +  \operatorname{Conv_2}(\operatorname{RB_2}(\operatorname{RB_1}(\operatorname{Conv_1}(\mathbf{r}^{(k)})))). 
\end{equation}

\subsection{Memory-Augmented DUN}
Motivated by the viewpoint that the human brain achieves memory by preserving and storing what they acquire or are informed previously \cite{cichon2015branch}, the information flow across stages can also be regarded as short-term memory transmission in DUNs, where $\mathbf{x}^{(k)}$ in Figure~\ref{fig:stage} is as the link between each two adjacent stages. However, $\operatorname{Conv_2}(\cdot)$ in Eq.~\eqref{eq:PMM} is a lossy conversion process which would cause memory loss. Here, we propose a \textbf{M}emory-\textbf{A}ugmented \textbf{D}eep \textbf{U}nfolding \textbf{N}etwork (\textbf{MADUN}) which augments memory across stages, each of which shares a unified two-step recovering scheme as follows:
\begin{numcases}{}
\mathbf{r}^{(k)}= \displaystyle{\mathcal{H}}_{\scriptscriptstyle{\operatorname{GDM}}}^{(k)}(\mathbf{x}^{(k-1)}),\\
\mathbf{x}^{(k)}=\displaystyle{\mathcal{H}}_{\scriptscriptstyle{\operatorname{MAPMM}}}^{(k)}(\mathbf{r}^{(k)}).
\end{numcases}
Here, GDM is trivial \cite{zhang2018ista} and is the same as Eq.~\eqref{eq:r}, and \textbf{M}emory-\textbf{A}ugmented \textbf{P}roximal-\textbf{M}apping \textbf{M}odule (\textbf{MAPMM}) is introduced with two types of memory augmentation mechanisms which can highly strengthen the information transmission among all stages.

\subsection{Memory-Augmented Proximal Mapping}
On the basis of PMM, MAPMM introduces \textbf{H}igh-throughput \textbf{S}hort-term \textbf{M}emory (\textbf{HSM}) and \textbf{C}ross-stage \textbf{L}ong-term \textbf{M}emory (\textbf{CLM}), which greatly enhance the network representation capability.

\subsubsection{\textbf{High-throughput Short-term Memory (HSM)}}
To address the information-lossy intra-stage transmission, we introduce HSM with multi-channel $\mathbf{z}^{(k-1)}$ (of size $H\times W \times C$) to bridge a high-capacity path between adjacent stages.
As shown in Figure~\ref{fig:stage} (c) and as the output of $\operatorname{RB_2}(\cdot)$ at $(k-1)$-th stage, $\mathbf{z}^{(k-1)}$ is: 
\begin{equation}
\mathbf{z}^{(k-1)} = \operatorname{RB_2}(\mathbf{s}^{(k-1)}). 
\end{equation}
Concatenating $\mathbf{z}^{(k-1)}$ with $\mathbf{r}^{(k)}$ followed by $\operatorname{Conv_3}(\cdot)$ generates $C$-channel features. The remaining process is  the same with Eq.~\eqref{eq:PMM}. Thus, PMM with HSM, denoted by $\mathbf{x}^{(k)}=\displaystyle{\mathcal{H}}_{\scriptscriptstyle{\operatorname{PMM-HSM}}}^{(k)}(\mathbf{r}^{(k)})$, is: 
\begin{equation}
\mathbf{x}^{(k)} = \mathbf{r}^{(k)} +  \operatorname{Conv_2}(\operatorname{RB_2}(\operatorname{RB_1}(\operatorname{Conv_3}(\mathbf{r}^{(k)} \| \mathbf{z}^{(k-1)})))), 
\end{equation}
where $\|$ denotes concatenating the feature maps. Obviously, $\mathbf{z}^{(k-1)}$ transmits high-throughput information from the previous stage to the current stage, achieving multi-channel short-term memory which ensures maximum flow and reduces information loss caused by the conventional channel-shrinking transformation.

\subsubsection{\textbf{Cross-stage Long-term Memory (CLM)}}
For modeling explicitly the long-range dependencies among all cascading stages, we utilize a ConvLSTM layer \cite{shi2015convolutional}, which has been well-validated to be a stable and powerful way to balance the past and current states, to develop the cross-stage long-term memory (CLM) and further enhance the signal conversion and transmission.

 
 In the process of achieving CLM, the ConvLSTM layer is sandwiched between two residual blocks ($\operatorname{RB_1}(\cdot)$ and $\operatorname{RB_2}(\cdot)$), and updates cell outputs $\mathbf{c}^{(k)}$ and hidden states $\mathbf{h}^{(k)}$ with the inputs of the intermediate result $\mathbf{s}^{(k)}=\operatorname{RB_1}(\operatorname{Conv_1}(\mathbf{r}^{(k)}))$ in Figure~\ref{fig:stage} (d),
\begin{equation}
[\mathbf{h}^{(k)}, \mathbf{c}^{(k)}] = \operatorname{ConvLSTM}(\mathbf{s}^{(k)}, [\mathbf{h}^{(k-1)}, \mathbf{c}^{(k-1)}]), 
\end{equation}
where CLM $[\mathbf{h}^{(k-1)}$, $\mathbf{c}^{(k-1)}]$ updates the current state of deep information, predicts multi-stage information and achieves long-term memory across cascading stages.
The architecture of ConvLSTM with inputs $\mathbf{s}^{(k)}$ can be expressed as:
\begin{equation}
\begin{aligned}
\mathbf{i}^{(k)} &=\sigma({\emph{W}_{si}} \ast \mathbf{s}^{(k)} + {\emph{W}_{hi}} \ast \mathbf{h}^{(k-1)} + \emph{b}_i), \\
\mathbf{f}^{(k)} &=\sigma({\emph{W}_{sf}} \ast \mathbf{s}^{(k)} + {\emph{W}_{hf}} \ast \mathbf{h}^{(k-1)} + \emph{b}_f), \\
\mathbf{c}^{(k)} &=\mathbf{f}^{(k)} \circ \mathbf{c}^{(k-1)} + \mathbf{i}^{(k)} \circ \operatorname{tanh}({\emph{W}_{sc}} \ast \mathbf{s}^{(k)} + {\emph{W}_{hc}} \ast \mathbf{h}^{(k-1)} + \emph{b}_c), \\
\mathbf{o}^{(k)} &=\sigma({\emph{W}_{so}} \ast \mathbf{s}^{(k)} + {\emph{W}_{ho}} \ast \mathbf{h}^{(k-1)} + \emph{b}_o), \\
\mathbf{h}^{(k)} &=\mathbf{o}^{(k)} \circ \operatorname{tanh}(\mathbf{c}^{(k)}), 
\end{aligned}
\end{equation}
where $\ast$ denotes the convolution operator, $\circ$  denotes the Hadamard product, $\emph{W}_{si}, \emph{W}_{hi}, \ldots, \emph{W}_{ho}$ are the filter weights, $\emph{b}_i, \emph{b}_f, \ldots, \emph{b}_o$ are the bias, $\mathbf{i}^{(k)}$, $\mathbf{f}^{(k)}, \mathbf{o}^{(k)}$ denote the input gate, the forget gate and the output gate respectively, $\sigma(\cdot)$ and $\operatorname{tanh}(\cdot)$ denote the sigmoid function and the tanh function respectively. $\mathbf{c}^{(k)}$ acts as an accumulator of the state information and $\mathbf{h}^{(k)}$ is further controlled by the lastest cell outputs $\mathbf{c}^{(k)}$ and the output gate $\mathbf{o}^{(k)}$.

 $\mathbf{h}^{(k)}$ is directly utilized as the inputs of $\operatorname{RB_2}(\cdot)$ in the current stage and $[\mathbf{h}^{(k)}$, $\mathbf{c}^{(k)}]$ transmits deep features at the same position across stages to augment the dependency of deep information. And the rest process of achieving CLM in PMM is the same with Eq. (7). So the process of $\mathbf{x}^{(k)}=\displaystyle{\mathcal{H}}_{\scriptscriptstyle{\operatorname{PMM-CLM}}}^{(k)}(\mathbf{r}^{(k)})$ can be reformulated as:
\begin{equation}
\mathbf{x}^{(k)} = \mathbf{r}^{(k)} +  \operatorname{Conv_2}(\operatorname{RB_2}(\operatorname{ConvLSTM}(\operatorname{RB_1}(\operatorname{Conv_1}(\mathbf{r}^{(k)}))))). 
\end{equation}

Therefore, incorporating both HSM $\mathbf{z}^{(k-1)}$ and CLM $[\mathbf{h}^{(k-1)},\\ \mathbf{c}^{(k-1)}]$ into PMM and with the inputs $\mathbf{r}^{(k)}$ in the ($k$)-th stage, the proposed memory-augmented proximal mapping module (MAPMM) in Figure~\ref{fig:stage} (e), denoted by  $\mathbf{x}^{(k)}=\displaystyle{\mathcal{H}}_{\scriptscriptstyle{\operatorname{MAPMM}}}^{(k)}(\mathbf{r}^{(k)})$ is formulated as:
\begin{equation}
\mathbf{x}^{(k)} = \mathbf{r}^{(k)} +  \operatorname{Conv_2}(\operatorname{RB_2}(\operatorname{ConvLSTM}(\operatorname{RB_1}(\operatorname{Conv_3}(\mathbf{r}^{(k)} \| \mathbf{z}^{(k-1)}))))). 
\label{eq:MAPMM}
\end{equation}
Comparing Eq.~\eqref{eq:MAPMM} with Eq.~\eqref{eq:PMM}, it is obvious to observe that $\mathbf{z}^{(k-1)}$ is able to effectively add high-throughput information between each two adjacent stages, while $[\mathbf{h}^{(k-1)}, \mathbf{c}^{(k-1)}]$ can augment mid/\\high-frequency information by applying  ConvLSTM layers across cascading stages. Therefore, our proposed MAPMM enhances information transmission among all stages by adaptively learning these two types of memory mechanisms.

\subsection{Initialization of HSM and CLM}
 Given the measurements $\mathbf{y}$ as the known information, HSM $\mathbf{z}^{(k)}$ is initialized by applying a light-weight one-convolution layer $\operatorname{Conv_0}(\cdot)$ on the image $\mathbf{\Phi^{\top}}\mathbf{y}$ to generate $C$ feature maps, yielding:
\begin{equation}
\mathbf{z}^{(0)} = \operatorname{Conv_0}(\mathbf{\Phi^{\top}}\mathbf{y}). 
\end{equation}
Considering that CLM $[\mathbf{h}^{(k)}$, $\mathbf{c}^{(k)}]$ has its physical meaning and it is practical to assume no prior knowledge is used at the beginning, therefore, we simply initialize $[\mathbf{h}^{(k)}$, $\mathbf{c}^{(k)}]$ to be zero. 

\subsection{Network Parameters and Loss Function}

The learnable parameter set in MADUN, denoted by $\boldsymbol{\Theta}$, can be expressed as 
 $\boldsymbol{\Theta}=\{\operatorname{Conv_0}(\cdot)\}\bigcup\{\rho^{(k)}, \displaystyle{\mathcal{H}}_{\scriptscriptstyle{\operatorname{MAPMM}}}^{(k)}(\cdot)\}_{k=1}^{K}$.
 $\operatorname{Conv_0}(\cdot)$ is a one-convolution layer with one input channel and $C$ output channels. $\displaystyle{\mathcal{H}}_{\scriptscriptstyle{\operatorname{MAPMM}}}^{(k)}(\cdot)$ consists of $\operatorname{Conv_3}(\cdot)$, $\operatorname{Conv_2}(\cdot)$, $\operatorname{RB_1}(\cdot)$, $\operatorname{RB_2}(\cdot)$ and $\operatorname{ConvLSTM}(\cdot)$ whose parameters are different in each stage. $\operatorname{Conv_3}(\cdot)$ with $C+1$ input channels and $C$ output channels receives networks inputs, and $\operatorname{Conv_2}(\cdot)$ with $C$ input channels and one output channel outputs residual results. Also, 
 the parameters of $\operatorname{RB_1}(\cdot)$ and $\operatorname{RB_2}(\cdot)$ are both from two convolution layers without the change of the channel number (with $C$ input channels and $C$ output channels), while $\operatorname{ConvLSTM}(\cdot)$ is same with \cite{li2018recurrent}. Here, all the convolutions in MADUN adopt $3\times3$ filter kernels.
\begin{figure}[t]
\setlength{\abovecaptionskip}{5pt}
\setlength{\belowcaptionskip}{0pt}
\centering
\includegraphics[width=0.9\linewidth]{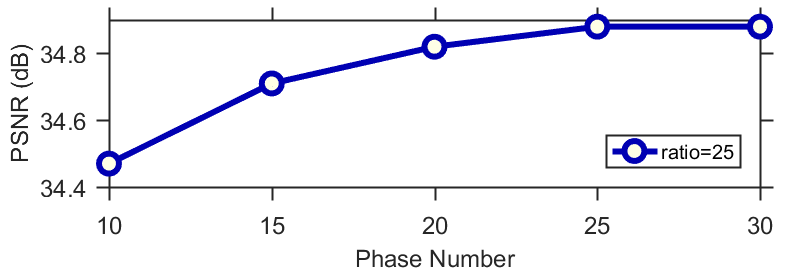} 
\caption{Average PSNR curve by MADUN on Set11 dataset with various stage numbers. When stage number is 25, the curve begins to reach peak performance.}
\vspace{-6pt}
\label{fig:stage_num}
\end{figure}
\begin{table}[t]
\centering
\small
\setlength{\abovecaptionskip}{5pt}
\setlength{\belowcaptionskip}{0pt}
\caption{Ablation study on effects of components at ratio=$10\%$. $\ast$HSM and $\circ$HSM represent the different positions of linking memory. The best performance is labeled in bold.} 
\label{tab:ablation study}
\renewcommand\tabcolsep{4pt}
\begin{tabular}{ccccccc}
\toprule[1pt]
\multicolumn{1}{c|}{\multirow{2}{*}{Cases}} & \multicolumn{1}{c}{\multirow{2}{*}{$\ast$HSM}} &
\multicolumn{1}{c}{\multirow{2}{*}{$\circ$HSM}} & \multicolumn{1}{c|}{\multirow{2}{*}{HSM}} & \multicolumn{1}{c|}{\multirow{2}{*}{CLM}} & \multicolumn{2}{c}{PSNR/SSIM} \\ \cline{6-7}  \multicolumn{1}{c|}{} & \multicolumn{1}{c}{} & \multicolumn{1}{c}{} & \multicolumn{1}{c|}{} & \multicolumn{1}{c|}{}
& \multicolumn{1}{c|}{Set11}         & \multicolumn{1}{c}{Urban100}
\\ \hline  
\multicolumn{1}{c|}{(a)} & \multicolumn{1}{c}{-} & \multicolumn{1}{c}{-} & \multicolumn{1}{c|}{-} & \multicolumn{1}{c|}{-} & \multicolumn{1}{r|}{28.66/0.8609} & \multicolumn{1}{c}{25.77/0.7905} 
\\
\multicolumn{1}{c|}{(b)} & \multicolumn{1}{c}{-} & \multicolumn{1}{c}{-} & \multicolumn{1}{c|}{-} & \multicolumn{1}{c|}{$\surd$} & \multicolumn{1}{r|}{29.24/0.8760} & \multicolumn{1}{c}{26.68/0.8282} 
\\ 
\multicolumn{1}{c|}{(c)} & \multicolumn{1}{c}{$\surd$} & \multicolumn{1}{c}{-} & \multicolumn{1}{c|}{-} & \multicolumn{1}{c|}{-} & \multicolumn{1}{r|}{29.06/0.8713} & \multicolumn{1}{c}{26.46/0.8207} 
\\ 
\multicolumn{1}{c|}{(d)} & \multicolumn{1}{c}{-} & \multicolumn{1}{c}{$\surd$} & \multicolumn{1}{c|}{-} & \multicolumn{1}{c|}{-} & \multicolumn{1}{r|}{29.32/0.8770} & \multicolumn{1}{c}{26.88/0.8263} 
\\ 
\multicolumn{1}{c|}{(e)} & \multicolumn{1}{c}{-} & \multicolumn{1}{c}{-} & \multicolumn{1}{c|}{$\surd$} & \multicolumn{1}{c|}{-} & \multicolumn{1}{r|}{29.35/0.8786} & \multicolumn{1}{c}{26.94/0.8305} 
\\
\multicolumn{1}{c|}{(f)} & \multicolumn{1}{c}{-} & \multicolumn{1}{c}{-} & \multicolumn{1}{c|}{$\surd$} & \multicolumn{1}{c|}{$\surd$} & \multicolumn{1}{r|}{\bf 29.44/0.8807} & \multicolumn{1}{c}{\bf 27.13/0.8393} 
\\ \toprule[1pt]
\end{tabular}
\end{table}

Given a set of full-sampled images $\left\{\mathbf{x}_j\right\}_{j=1}^{N_a}$ and some sampling patterns with CS ratio $\gamma$, the under-sampled $k$-space data is obtained by $\mathbf{y}_j=\mathbf{\Phi}\mathbf{x}_j$, producing the train data pairs $\{(\mathbf{y}_j, \mathbf{x}_j)\}_{j=1}^{N_a}$. Our MADUN takes $\mathbf{y}_j$ as inputs and generates the reconstruction result $\mathbf{x}_j^{(K)}$ as outputs with the initialization $\mathbf{x}_j^{(0)}=\mathbf{\Phi^{\top}}\mathbf{y}_j$. The loss function is designed to use $L_1$ loss between $\mathbf{x}_j$ and $\mathbf{x}_j^{(K)}$ as:
\begin{equation} 
\setlength{\abovedisplayskip}{5pt}
\setlength{\belowdisplayskip}{6pt}
\label{eq: fifteen}
\mathcal{L}(\boldsymbol{\Theta}) =\ \frac{1}{{N}{N_a}}\sum_{j=1}^{N_a} {\left\|\mathbf{x}_j-\mathbf{x}_j^{(K)}\right\|}_1, 
\end{equation}
where $N_a$, $N$ and $K$ represent the number of training images, the size of each image and the stage number of MADUN respectively.

\section{experiment}
\subsection{Implementation Details}
\label{ssec:Implementation Details}
We use the 400 training images of size $180 \times 180$ \cite{chen2016trainable}, generating the training data pairs $\{(\mathbf{y}_j, \mathbf{x}_j)\}_{j=1}^{N_a}$ by extracting the luminance component of each image block of size $33 \times 33$,
\textit{i.e.} $N=1,089$. Meanwhile, we apply in the data augmentation technique to increase the data diversity. For a given CS ratio, the corresponding measurement matrix $\mathbf{\Phi}\in\mathbb{R}^{M \times N}$ is constructed by generating a random Gaussian matrix or a jointly learned matrix and then orthogonalizing its rows, \textit{i.e.} $\mathbf{\Phi}\mathbf{\Phi^{\top}}=\mathbf{I}$, where $\mathbf{I}$ is the identity matrix. Applying $\mathbf{y}_j=\mathbf{\Phi}\mathbf{x}_j$ yields the set of CS measurements, where $\mathbf{x}_j$ is the vectorized version of an image block. 

\begin{figure}[t]
\centering
\setlength{\abovecaptionskip}{5pt}
\setlength{\belowcaptionskip}{0pt}
\includegraphics[width=1\linewidth]{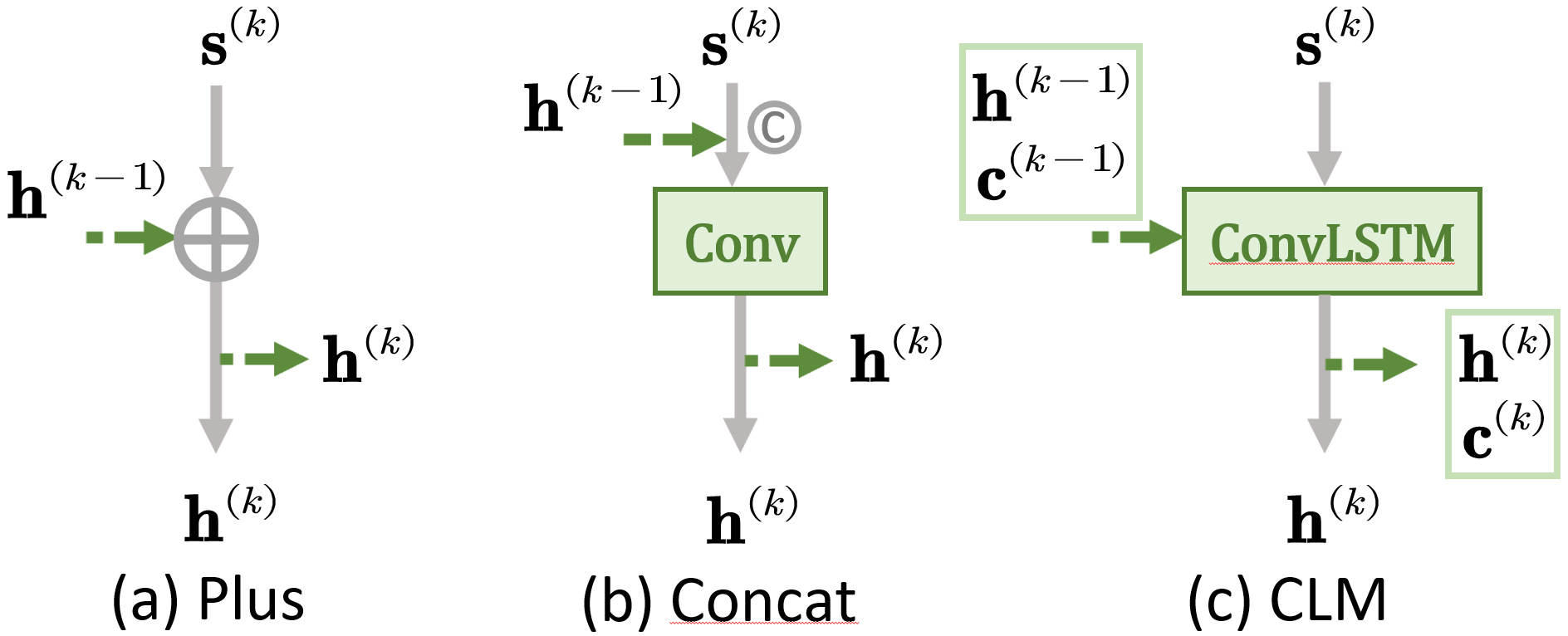} 
\caption{Illustration of three different methods achieving memory across stages, corresponding to Table~\ref{tab:CLM}. 
}
\label{fig:CLM}
\vspace{-6pt}
\end{figure}
\begin{table}[t]
\centering
\setlength{\abovecaptionskip}{5pt}
\setlength{\belowcaptionskip}{0pt}
\caption{Ablation study on the effect of CLM. Average PSNR/SSIM at ratio=$10\%, 25\%, 30\%$ on Set11 dataset. The best performance is labeled in bold.} 
\label{tab:CLM}
\begin{tabular}{ccccc}
\toprule[1pt]
\multicolumn{1}{c|}{\multirow{1}{*}{Methods}} & \multicolumn{1}{c|}{\multirow{1}{*}{Plus}}      & \multicolumn{1}{c|}{\multirow{1}{*}{Concat}} & \multicolumn{1}{c}{\multirow{1}{*}{CLM}} 
\\ \hline  
\multicolumn{1}{c|}{10\%}  & \multicolumn{1}{c|}{28.83/0.8659} & \multicolumn{1}{c|}{29.10/0.8733} & \multicolumn{1}{c}{\bf 29.24/0.8760} 
\\
\multicolumn{1}{c|}{25\%} & \multicolumn{1}{c|}{34.33/0.9446} & \multicolumn{1}{c|}{34.54/0.9465} & \multicolumn{1}{c}{\bf 34.69/0.9480} 
\\ 
\multicolumn{1}{c|}{30\%} & \multicolumn{1}{c|}{35.62/0.9547} & \multicolumn{1}{c|}{35.74/0.9560} & \multicolumn{1}{c}{\bf 35.93/0.9570} 
\\ \toprule[1pt]
\end{tabular}
\end{table}
\begin{figure}[t]
\centering
\setlength{\abovecaptionskip}{5pt}
\setlength{\belowcaptionskip}{0pt}
\includegraphics[width=1\linewidth]{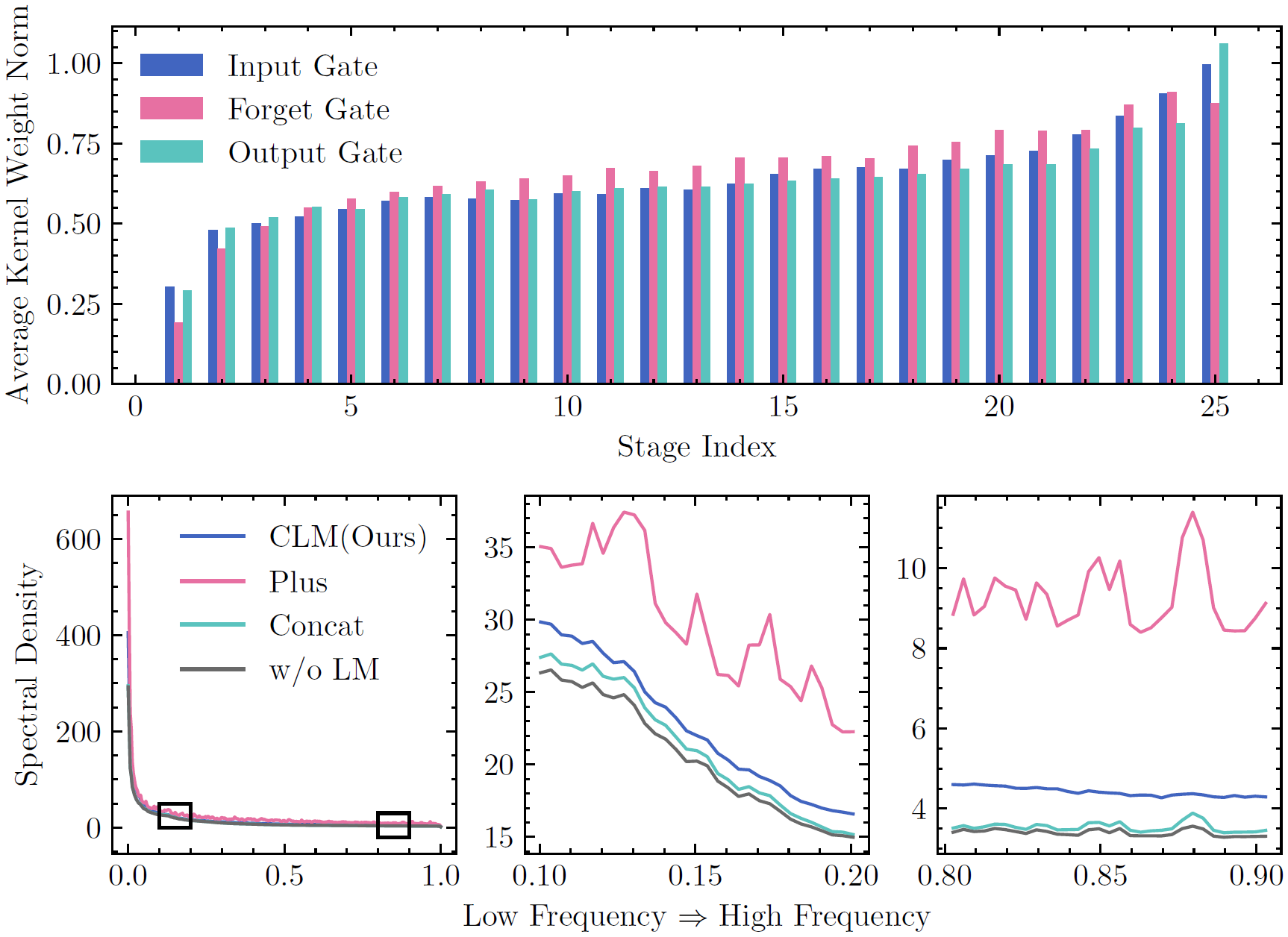} 
\caption{Illustration of the average kernel weight norm in each stage (upper) and the 1-D spectral density curves of different CLM variants (lower). 
}
\label{fig:CLM_analysis}
\vspace{-6pt}
\end{figure}
\begin{table*}[!htbp]
\centering
\setlength{\abovecaptionskip}{5pt}
\setlength{\belowcaptionskip}{0pt}
\small
\caption{Average PSNR/SSIM performance comparisons on Set11, CBSD68 and Urban100 datasets with different CS ratios. We compare MADUN with two DNUNs and six DUNs with fixed random gaussian matrix. The best and second best results are highlighted in \textcolor{red}{red} and \textcolor{blue}{blue} colors, respectively.} 
\label{tab:fixed_matrix}
\renewcommand\tabcolsep{13pt}
\begin{tabular}{c|c|ccccc}
\toprule[1pt]
\multicolumn{1}{c|}{\multirow{2}{*}{Dataset}}  &
\multicolumn{1}{c|}{\multirow{2}{*}{Methods}}   & 
\multicolumn{5}{c}{CS Ratio} 
\\ \cline{3-7}  
\multicolumn{1}{c|}{} &  \multicolumn{1}{c|}{}  &
\multicolumn{1}{c}{10\%}  & \multicolumn{1}{c}{25\%}  & \multicolumn{1}{c}{30\%}         
& \multicolumn{1}{c}{40\%}  & \multicolumn{1}{c}{50\%} \\
\toprule[1pt]
\multicolumn{1}{c|}{}  &
\multicolumn{1}{l|}{ReconNet \cite{Kulkarni2016ReconNetNR}}      & \multicolumn{1}{c}{23.96/0.7172} &
\multicolumn{1}{c}{26.38/0.7883} & \multicolumn{1}{c}{28.20/0.8424} & 
\multicolumn{1}{c}{30.02/0.8837} & \multicolumn{1}{c}{30.62/0.8983} \\
\multicolumn{1}{c|}{}  &
\multicolumn{1}{l|}{DPA-Net \cite{sun2020dual}}      & \multicolumn{1}{c}{27.66/0.8530} &
\multicolumn{1}{c}{32.38/0.9311} & \multicolumn{1}{c}{33.35/0.9425} & 
\multicolumn{1}{c}{35.21/0.9580} & \multicolumn{1}{c}{36.80/0.9685} \\
\cline{2-7}
\multicolumn{1}{c|}{}  &
\multicolumn{1}{l|}{IRCNN \cite{zhang2017learning}}      & \multicolumn{1}{c}{23.05/0.6789} &
\multicolumn{1}{c}{28.42/0.8382} & \multicolumn{1}{c}{29.55/0.8606} & 
\multicolumn{1}{c}{31.30/0.8898} & \multicolumn{1}{c}{32.59/0.9075} \\
\multicolumn{1}{c|}{}  &
\multicolumn{1}{l|}{ISTA-Net$^+$ \cite{zhang2018ista}}      & \multicolumn{1}{c}{26.58/0.8066} &
\multicolumn{1}{c}{32.48/0.9242} & \multicolumn{1}{c}{33.81/0.9393} & 
\multicolumn{1}{c}{36.04/0.9581} & \multicolumn{1}{c}{38.06/0.9706} \\
\multicolumn{1}{c|}{}  &
\multicolumn{1}{l|}{DPDNN \cite{dong2018denoising}}      & \multicolumn{1}{c}{26.23/0.7992} &
\multicolumn{1}{c}{31.71/0.9153} & \multicolumn{1}{c}{33.16/0.9338} & 
\multicolumn{1}{c}{35.29/0.9534} & \multicolumn{1}{c}{37.63/0.9693} \\
\multicolumn{1}{c|}{Set11}  &
\multicolumn{1}{l|}{GDN \cite{gilton2019neumann}}      & \multicolumn{1}{c}{23.90/0.6927} &
\multicolumn{1}{c}{29.20/0.8600} & \multicolumn{1}{c}{30.26/0.8833} & 
\multicolumn{1}{c}{32.31/0.9137} & \multicolumn{1}{c}{33.31/0.9285} \\
\multicolumn{1}{c|}{}  &
\multicolumn{1}{l|}{MAC-Net \cite{chenlearning}}      & \multicolumn{1}{c}{27.68/0.8182} &
\multicolumn{1}{c}{32.91/0.9244} & \multicolumn{1}{c}{33.96/0.9372} & 
\multicolumn{1}{c}{35.94/0.9560} & \multicolumn{1}{c}{37.67/0.9668} \\
\multicolumn{1}{c|}{}  &
\multicolumn{1}{l|}{iPiano-Net \cite{su2020ipiano}}      & \multicolumn{1}{c}{\textcolor{blue}{28.05/0.8460}} &
\multicolumn{1}{c}{\textcolor{blue}{33.53/0.9359}} & \multicolumn{1}{c}{\textcolor{blue}{34.78/0.9472}} & 
\multicolumn{1}{c}{\textcolor{blue}{37.01/0.9631}} & \multicolumn{1}{c}{\textcolor{blue}{38.94/0.9737}} \\
\Xcline{2-7}{0.7pt}
\multicolumn{1}{c|}{}  &
\multicolumn{1}{l|}{MADUN}      & \multicolumn{1}{c}{\textcolor{red}{29.44/0.8807}} &
\multicolumn{1}{c}{\textcolor{red}{34.88/0.9496}} & \multicolumn{1}{c}{\textcolor{red}{36.07/0.9582}} & 
\multicolumn{1}{c}{\textcolor{red}{38.13/0.9700}} & \multicolumn{1}{c}{\textcolor{red}{39.92/0.9779}} \\
\toprule[1pt]
\multicolumn{1}{c|}{}  &
\multicolumn{1}{l|}{ReconNet \cite{Kulkarni2016ReconNetNR}}      & \multicolumn{1}{c}{24.02/0.6414} &
\multicolumn{1}{c}{26.01/0.7498} & \multicolumn{1}{c}{27.20/0.7909} & 
\multicolumn{1}{c}{28.71/0.8409} & \multicolumn{1}{c}{29.32/0.8642} \\
\multicolumn{1}{c|}{}  &
\multicolumn{1}{l|}{DPA-Net \cite{sun2020dual}}      & \multicolumn{1}{c}{25.47/0.7372} &
\multicolumn{1}{c}{29.01/0.8595} & \multicolumn{1}{c}{29.73/0.8827} & 
\multicolumn{1}{c}{31.17/0.9156} & \multicolumn{1}{c}{32.55/0.9386} \\
\cline{2-7}
\multicolumn{1}{c|}{}  &
\multicolumn{1}{l|}{IRCNN \cite{zhang2017learning}}      & \multicolumn{1}{c}{23.07/0.5580} &
\multicolumn{1}{c}{26.44/0.7206} & \multicolumn{1}{c}{27.31/0.7543} & 
\multicolumn{1}{c}{28.76/0.8042} & \multicolumn{1}{c}{30.00/0.8398} \\
\multicolumn{1}{c|}{}  &
\multicolumn{1}{l|}{ISTA-Net$^+$ \cite{zhang2018ista}}      & \multicolumn{1}{c}{25.37/0.7022} &
\multicolumn{1}{c}{29.32/0.8515} & \multicolumn{1}{c}{30.37/0.8786} & 
\multicolumn{1}{c}{32.23/0.9165} & \multicolumn{1}{c}{34.04/0.9425} \\
\multicolumn{1}{c|}{}  &
\multicolumn{1}{l|}{DPDNN \cite{dong2018denoising}}      & \multicolumn{1}{c}{25.35/0.7020} &
\multicolumn{1}{c}{29.28/0.8513} & \multicolumn{1}{c}{30.39/0.8807} & 
\multicolumn{1}{c}{32.21/0.9171} & \multicolumn{1}{c}{34.27/0.9455} \\
\multicolumn{1}{c|}{CBSD68}  &
\multicolumn{1}{l|}{GDN \cite{gilton2019neumann}}      & \multicolumn{1}{c}{23.41/0.6011} &
\multicolumn{1}{c}{27.11/0.7636} & \multicolumn{1}{c}{27.52/0.7745} & 
\multicolumn{1}{c}{30.14/0.8649} & \multicolumn{1}{c}{30.88/0.8763} \\
\multicolumn{1}{c|}{}  &
\multicolumn{1}{l|}{MAC-Net \cite{chenlearning}}      & \multicolumn{1}{c}{25.80/0.7024} &
\multicolumn{1}{c}{29.42/0.8469} & \multicolumn{1}{c}{30.28/0.8713} & 
\multicolumn{1}{c}{32.02/0.9085} & \multicolumn{1}{c}{33.68/0.9352} \\
\multicolumn{1}{c|}{}  &
\multicolumn{1}{l|}{iPiano-Net \cite{su2020ipiano}}      & \multicolumn{1}{c}{\textcolor{blue}{26.34/0.7431}} &
\multicolumn{1}{c}{\textcolor{blue}{30.16/0.8711}} & \multicolumn{1}{c}{\textcolor{blue}{31.24/0.8964}} & 
\multicolumn{1}{c}{\textcolor{blue}{33.14/0.9298}} & \multicolumn{1}{c}{\textcolor{blue}{34.98/0.9521}} \\
\Xcline{2-7}{0.7pt}
\multicolumn{1}{c|}{}  &
\multicolumn{1}{l|}{MADUN}      & \multicolumn{1}{c}{\textcolor{red}{26.83/0.7620}} &
\multicolumn{1}{c}{\textcolor{red}{30.81/0.8844}} & \multicolumn{1}{c}{\textcolor{red}{31.87/0.9068}} & 
\multicolumn{1}{c}{\textcolor{red}{33.81/0.9376}} & \multicolumn{1}{c}{\textcolor{red}{35.82/0.9587}} \\
\toprule[1pt]
\multicolumn{1}{c|}{}  &
\multicolumn{1}{l|}{ReconNet \cite{Kulkarni2016ReconNetNR}}      & \multicolumn{1}{c}{21.49/0.6223} &
\multicolumn{1}{c}{23.31/0.7107} & \multicolumn{1}{c}{24.72/0.7697} & 
\multicolumn{1}{c}{26.44/0.8250} & \multicolumn{1}{c}{27.06/0.8447} \\
\multicolumn{1}{c|}{}  &
\multicolumn{1}{l|}{DPA-Net \cite{sun2020dual}}      & \multicolumn{1}{c}{24.55/0.7841} &
\multicolumn{1}{c}{28.80/0.8944} & \multicolumn{1}{c}{29.47/0.9034} & 
\multicolumn{1}{c}{31.09/0.9311} & \multicolumn{1}{c}{32.08/0.9447} \\
\cline{2-7}
\multicolumn{1}{c|}{}  &
\multicolumn{1}{l|}{IRCNN \cite{zhang2017learning}}      & \multicolumn{1}{c}{21.62/0.6137} &
\multicolumn{1}{c}{26.38/0.7955} & \multicolumn{1}{c}{27.47/0.8248} & 
\multicolumn{1}{c}{29.22/0.8645} & \multicolumn{1}{c}{30.63/0.8900} \\
\multicolumn{1}{c|}{}  &
\multicolumn{1}{l|}{ISTA-Net$^+$ \cite{zhang2018ista}}      & \multicolumn{1}{c}{23.61/0.7238} &
\multicolumn{1}{c}{28.93/0.8840} & \multicolumn{1}{c}{30.21/0.9079} & 
\multicolumn{1}{c}{32.43/0.9377} & \multicolumn{1}{c}{34.43/0.9571} \\
\multicolumn{1}{c|}{}  &
\multicolumn{1}{l|}{DPDNN \cite{dong2018denoising}}      & \multicolumn{1}{c}{23.69/0.7211} &
\multicolumn{1}{c}{28.70/0.8798} & \multicolumn{1}{c}{30.06/0.9005} & 
\multicolumn{1}{c}{32.27/0.9350} & \multicolumn{1}{c}{34.81/0.9579} \\
\multicolumn{1}{c|}{Urban100}  &
\multicolumn{1}{l|}{GDN \cite{gilton2019neumann}}      & \multicolumn{1}{c}{21.48/0.5958} &
\multicolumn{1}{c}{25.75/0.7704} & \multicolumn{1}{c}{26.72/0.7950} & 
\multicolumn{1}{c}{28.96/0.8698} & \multicolumn{1}{c}{29.89/0.8768} \\
\multicolumn{1}{c|}{}  &
\multicolumn{1}{l|}{MAC-Net \cite{chenlearning}}      & \multicolumn{1}{c}{24.21/0.7445} &
\multicolumn{1}{c}{28.79/0.8798} & \multicolumn{1}{c}{29.99/0.9017} & 
\multicolumn{1}{c}{31.94/0.9272} & \multicolumn{1}{c}{34.03/0.9513} \\
\multicolumn{1}{c|}{}  &
\multicolumn{1}{l|}{iPiano-Net \cite{su2020ipiano}}      & \multicolumn{1}{c}{{\textcolor{blue}{25.67/0.7963}}} &
\multicolumn{1}{c}{\textcolor{blue}{30.87/0.9157}} & \multicolumn{1}{c}{\textcolor{blue}{32.16/0.9320}} & 
\multicolumn{1}{c}{\textcolor{blue}{34.27/0.9531}} & \multicolumn{1}{c}{\textcolor{blue}{36.22/0.9675}} \\
\Xcline{2-7}{0.7pt}
\multicolumn{1}{c|}{}  &
\multicolumn{1}{l|}{MADUN}      & \multicolumn{1}{c}{\textcolor{red}{27.13/0.8393}} &
\multicolumn{1}{c}{\textcolor{red}{32.54/0.9347}} & \multicolumn{1}{c}{\textcolor{red}{33.77/0.9472}} & 
\multicolumn{1}{c}{\textcolor{red}{35.80/0.9633}} & \multicolumn{1}{c}{\textcolor{red}{37.75/0.9746}} \\
\toprule[1pt]
\end{tabular}
\end{table*}

To train the network, we use Adam optimization \cite{Kingma2015AdamAM} with a learning rate of 0.0001, and a batch size of 64. We also use momentum of 0.9 and weight decay of 0.999. 
For MADUN, the models are trained with 410 epochs separately for each CS ratio. Each image block of size $33 \times 33$ is sampled and reconstructed independently for the first 400 epochs, and for the last ten epochs, we adopt larger image blocks of size $99 \times 99$ as inputs to further fine-tune the model. To alleviate blocking artifacts, we firstly unfold the blocks of size $99 \times 99$ into overlapping blocks of size $33 \times 33$ while sampling process $\mathbf{\Phi}\mathbf{x}$ and then fold the blocks of size $33 \times 33$ into larger blocks while initialization $\mathbf{\Phi^\top}\mathbf{y}$ \cite{su2020ipiano}. We also unfold the whole image with this approach during testing. The default stage number $K$ is set to be 25 and the default number of feature maps $C$ is set to be 32. And the learnable parameters $\rho^{(k)}$ is initialized to 1. The CS reconstruction accuracies on the all datasets are evaluated with PSNR and SSIM.

\subsection{Ablation Study}

\subsubsection{\textbf{Performance Effect of Stage Number}}
The average PSNR curve in Figure~\ref{fig:stage_num} is the results when ratio is $25\%$ on Set11 dataset. 
To get a better trade-off between model performance and complexity, we choose $K=25$ as the default setting in all experiments.

\subsubsection{\textbf{Performance Effect of MADUN Components}}
Based on the PMM structure, we do ablation experiments on the HSM and CLM at ratio=$10\%$ on Set11 and Urban100 dataset, as shown in Table~\ref{tab:ablation study}. To demonstrate the effectiveness of HSM and CLM, we removes them from MADUN to form two simplified versions, corresponding to Figure~\ref{fig:stage} (c) and Figure~\ref{fig:stage} (d) respectively. We conclude that they are all effective ways to compensate the signal flow among stages, and HSM plays a more important role by bridging each two adjacent stages with a high-throughput short-term transmission.


\subsection{Analysis of Memory Augmentation}
\begin{figure*}[t]
\centering
\setlength{\abovecaptionskip}{5pt}
\setlength{\belowcaptionskip}{0pt}
\includegraphics[width=1\textwidth]{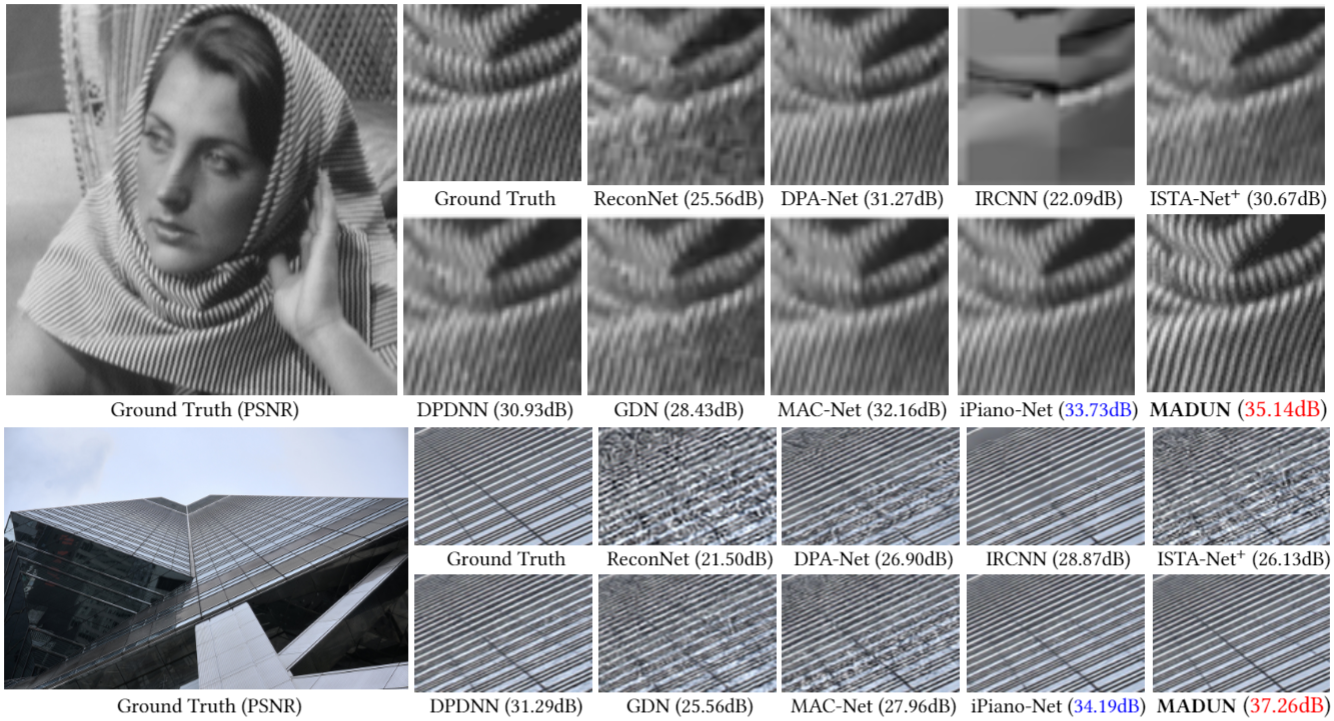} 
\caption{Comparisons on recovering an image named ``Barbara'' from Set11 dataset in the case of CS ratio = 30\% (upper) and an image from Urban100 dataset in the case of CS ratio = 40\% (lower).  
}
\label{fig:fixed_matrix}
\vspace{-5pt}
\end{figure*}

\begin{figure*}[t]
\centering
\setlength{\abovecaptionskip}{5pt}
\setlength{\belowcaptionskip}{0pt}
\includegraphics[width=1\textwidth]{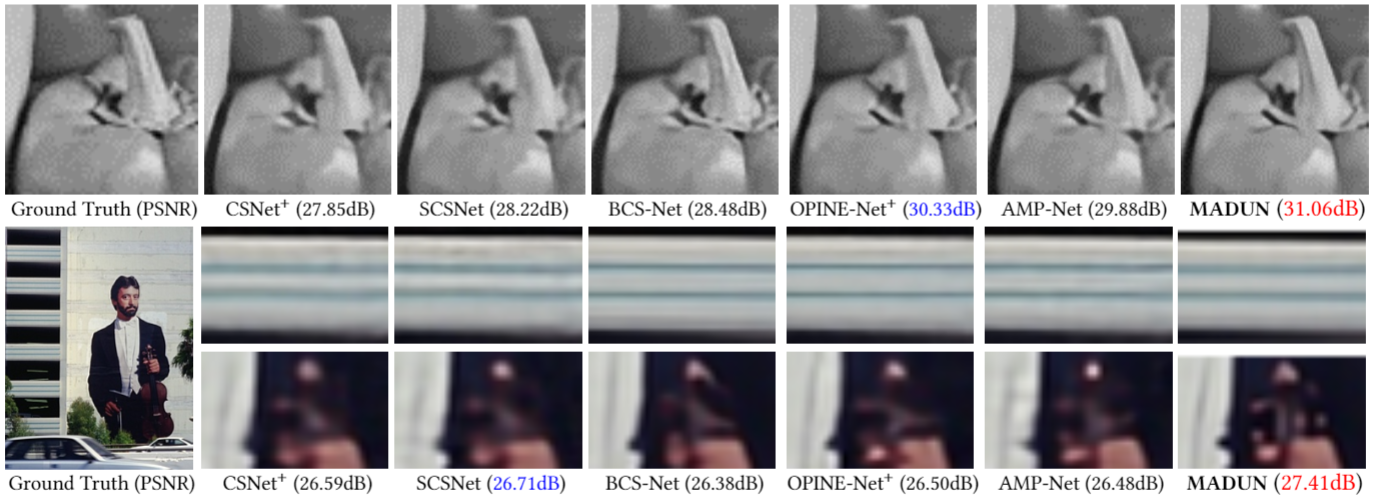} 
\caption{Comparisons on recovering an image named ``Peppers'' from Set11 dataset (upper) and an image from CBSD68 dataset (lower) in the case of CS ratio = 10\%.  
}
\label{fig:joint_matrix}
\vspace{-5pt}
\end{figure*}
\begin{table*}
		\centering
		\setlength{\abovecaptionskip}{5pt}
        \setlength{\belowcaptionskip}{0pt}
		\small
		\caption{Average PSNR/SSIM performance comparisons on Set11 and CBSD68 datasets with different CS ratios. We compare MADUN with two DNUNs and three DUNs with jontly learning the sampling matrix and the reconstruction process.}
		\label{tab:joint_matrix}
		\resizebox{0.92\textwidth}{!}{%
		\renewcommand\tabcolsep{13pt}
			\begin{tabular}{c|c|ccccc}
			\toprule[1pt]
				\multirow{2}{*}{Dataset}&\multirow{2}{*}{Methods}&\multicolumn{5}{c}{CS Ratio}\\ \cline{3-7}
				&&10\%&25\%&30\%&40\%&50\%\\
				\toprule[1pt]
				\multirow{5}{*}{Set11}
				&\multicolumn{1}{l|}{CSNet$^{+}$\cite{shi2019image}} &28.34/0.8580&33.34/0.9387&34.27/0.9492&36.44/0.9690&38.47/0.9796\\
				&\multicolumn{1}{l|}{SCSNet \cite{shi2019scalable}}
				&28.52/0.8616&33.43/0.9373&34.64/0.9511&36.92/0.9666&39.01/0.9769\\ \cline{2-7}
				&\multicolumn{1}{l|}{BCS-Net \cite{zhou2020multi}}
				&29.42/0.8673&34.20/0.9408&35.63/0.9495&36.68/0.9667&39.58/0.9734\\ 
				&\multicolumn{1}{l|}{OPINE-Net$^{+}$\cite{zhang2020optimization}}
				&\textcolor{blue}{29.81/0.8904}&\textcolor{blue}{34.86/0.9509}&35.79/0.9541&37.96/0.9633&40.19/\textcolor{blue}{0.9800}\\
				&\multicolumn{1}{l|}{AMP-Net \cite{zhang2020amp}}
				&29.42/0.8782&34.60/0.9469&\textcolor{blue}{35.91/0.9576}&\textcolor{blue}{38.25/0.9714}&\textcolor{blue}{40.26}/0.9786\\
				\Xcline{2-7}{0.7pt}
				&\multicolumn{1}{l|}{MADUN}
				&\textcolor{red}{29.91/0.8986}&\textcolor{red}{35.66/0.9601}&\textcolor{red}{36.94/0.9676}&\textcolor{red}{39.15/0.9772}&\textcolor{red}{40.77/0.9832}\\
				\toprule[1pt]
				\multirow{5}{*}{CBSD68}
				&\multicolumn{1}{l|}{CSNet$^{+}$\cite{shi2019image}}
				&27.91/0.7938&31.12/0.9060&32.20/0.9220&35.01/0.9258&36.76/0.9638\\
				&\multicolumn{1}{l|}{SCSNet \cite{shi2019scalable}}
				&\textcolor{blue}{28.02}/0.8042&31.15/0.9058&32.64/0.9237&35.03/0.9214&36.27/0.9593\\ \cline{2-7}
				&\multicolumn{1}{l|}{BCS-Net \cite{zhou2020multi}}
				&27.98/0.8015&31.29/0.8846&\textcolor{blue}{32.70/0.9301}&\textcolor{blue}{35.14/0.9397}&\textcolor{blue}{36.85/0.9682}\\ 
				&\multicolumn{1}{l|}{OPINE-Net$^{+}$\cite{zhang2020optimization}}
				&27.82/\textcolor{blue}{0.8045}&\textcolor{blue}{31.51/0.9061}&32.35/0.9215&34.95/0.9261&36.35/0.9660\\
				&\multicolumn{1}{l|}{AMP-Net \cite{zhang2020amp}}
				&27.79/0.7853&31.37/0.8749&32.68/0.9291&35.06/0.9395&36.59/0.9620\\
				\Xcline{2-7}{0.7pt}
				&\multicolumn{1}{l|}{MADUN}
				&\textcolor{red}{28.15/0.8229}&\textcolor{red}{32.26/0.9221}&\textcolor{red}{33.35/0.9379}&\textcolor{red}{35.42/0.9606}&\textcolor{red}{37.11/0.9730}\\
				\toprule[1pt]
		\end{tabular}}
	\end{table*}
\begin{table*}[t]
\centering
\setlength{\abovecaptionskip}{5pt}
\setlength{\belowcaptionskip}{0pt}
\small
\caption{Average PSNR/SSIM performance comparisons on testing brain MR images with one DNUN and six DUNs.} 
\label{tab:brain600}
\renewcommand\tabcolsep{5pt}
\begin{tabular}{ccccccccc}
\toprule[1pt]
\multicolumn{1}{c!{\vrule width1pt}}{CS Ratio}
&
\multicolumn{1}{c|}{Hyun et al.
\cite{Hyun_2018}
}     &
\multicolumn{1}{c}{Schlemper et al.
\cite{8067520}
}  &
\multicolumn{1}{c}{ADMM-Net
\cite{yang2018admm}
}     &
\multicolumn{1}{c}{RDN
\cite{Sun2018CompressedSM}
}   & 
\multicolumn{1}{c}{CDDN
\cite{Zheng2019CascadedDD}
}  &
\multicolumn{1}{c}{ISTA-Net$^+$
\cite{zhang2018ista}
} &
\multicolumn{1}{c!{\vrule width0.6pt}}{MoDL
\cite{aggarwal2018modl}
}     &
\multicolumn{1}{c}{MADUN} 
\\ \hline
\multicolumn{1}{c!{\vrule width1pt}}{10\%}         & 
\multicolumn{1}{c|}{32.78/0.8385} &
\multicolumn{1}{c}{34.23/0.8921} &
\multicolumn{1}{c}{34.42/0.8971} &
\multicolumn{1}{c}{34.59/0.8968} &
\multicolumn{1}{c}{34.63/0.9002} &
\multicolumn{1}{c}{34.65/0.9038} &
\multicolumn{1}{c!{\vrule width0.6pt}}{\textcolor{blue}{35.18/0.9091}} &
\multicolumn{1}{c}{\textcolor{red}{36.15/0.9237}} \\
\multicolumn{1}{c!{\vrule width1pt}}{20\%}         &
\multicolumn{1}{c|}{36.36/0.9070} &
\multicolumn{1}{c}{38.47/0.9457} &
\multicolumn{1}{c}{38.60/0.9478} &
\multicolumn{1}{c}{38.58/0.9470} &
\multicolumn{1}{c}{38.59/0.9474} &
\multicolumn{1}{c}{\textcolor{blue}{38.67/0.9480}} &
\multicolumn{1}{c!{\vrule width0.6pt}}{38.51/0.9457} &
\multicolumn{1}{c}{\textcolor{red}{39.44/0.9542}} \\
\multicolumn{1}{c!{\vrule width1pt}}{30\%}         &
\multicolumn{1}{c|}{38.85/0.9383} &
\multicolumn{1}{c}{40.85/0.9628} &
\multicolumn{1}{c}{40.87/0.9633} &
\multicolumn{1}{c}{40.82/0.9625} &
\multicolumn{1}{c}{40.89/0.9633} &
\multicolumn{1}{c}{40.91/0.9631} &
\multicolumn{1}{c!{\vrule width0.6pt}}{\textcolor{blue}{40.97/0.9636}} &
\multicolumn{1}{c}{\textcolor{red}{41.48/0.9666}} \\
\multicolumn{1}{c!{\vrule width1pt}}{40\%}         &
\multicolumn{1}{c|}{40.65/0.9539} &
\multicolumn{1}{c}{42.63/0.9724} &
\multicolumn{1}{c}{42.58/0.9726} &
\multicolumn{1}{c}{42.64/0.9723} &
\multicolumn{1}{c}{42.59/0.9725} &
\multicolumn{1}{c}{\textcolor{blue}{42.65/0.9727}} &
\multicolumn{1}{c!{\vrule width0.6pt}}{42.38/0.9705} & 
\multicolumn{1}{c}{\textcolor{red}{43.06/0.9746}} \\
\multicolumn{1}{c!{\vrule width1pt}}{50\%}          &
\multicolumn{1}{c|}{42.35/0.9662}  &
\multicolumn{1}{c}{44.19/0.9794}  &
\multicolumn{1}{c}{44.19/0.9796}  &
\multicolumn{1}{c}{44.18/0.9793}  &
\multicolumn{1}{c}{44.15/0.9795}  &
\multicolumn{1}{c}{\textcolor{blue}{44.24/0.9798}}  &
\multicolumn{1}{c!{\vrule width0.6pt}}{44.20/0.9776}  &
\multicolumn{1}{c}{\textcolor{red}{44.60/0.9810}} \\
\toprule[1pt]
\end{tabular}
\end{table*}
We now separately illustrate how our different memory augmentation mechanisms affect information transmission.

For HSM, high-throughput information which is added to the original one-channel short-memory between each two adjacent stages significantly reduces information loss and improves performance, as shown in Table~\ref{tab:ablation study}. For further analysis of the effect on different positions of linking memory in HSM, we make two comparative experiments (c) and (d), where $\ast$HSM represents that the outputs of $\operatorname{Conv_3}(\cdot)$ are the points of memory transmission and $\circ$HSM denotes the outputs of $\operatorname{RB_1}(\cdot)$. As we can see, our default version of HSM transmits more refined memory than others. 

And for CLM, we do experiments in different memory methods to prove the superiority of ConvLSTM, as shown in Table~\ref{tab:CLM}. In Figure~\ref{fig:CLM}, on the basis of PMM, (a) is the ``Plus'' strategy which takes a direct addition operation, (b) is the ``Concat'' strategy which adopts a single convolution layer to merge the all memory into the main intermediate features, and (c) is the implementation of CLM.
CLM achieves better performance than others, which reveals that information is selectively retained by applying LSTM is useful for CS. To get deeper insights of the long-term information flow established by our CLM, as Figure~\ref{fig:CLM_analysis} illustrates, we give the average kernel weight norms of the three convolution gates in each CLM, and plot the 1-D spectral density curves of memory features $\mathbf{h}^{(k)}$ in different CLM variants 
by integrating the 2-D power spectrums along each concentric circle \cite{tai2017memnet} (\textit{i.e.} by averaging the feature components with the same frequency in the DCT domain and normalizing the frequency range into [0,1]).
From the upper subplot, we observe that the weight norms of the three convolution gates increase as the stage index increases, and the average norm of the forget gate is larger than the other two in most stages, which means that the cell outputs $\mathbf{c}^{(k)}$ get close attention in the main trunk of MADUN and CLM plays a more important role in later stages. From the spectral density curves, we can see that, without long-term memory makes the network hard to keep the mid/high-frequency information; the ``Concat'' strategy also causes the information loss due to its weak adaptability; the ``Plus'' strategy introduces much feature space noise and weakens the proximal mapping process. However, our CLM with a flexible gated memory mechanism, achieves a better balance in the long-term information flow, which obtains the higher performances compared with others.

\subsection{Qualitative Evaluation}
We conduct two types of experiments on fixed random Gaussian sampling matrix and jointly learned sampling matrix, and compare them with the corresponding methods respectively.

\subsubsection{\textbf{Fixed Random Gaussian Matrix}}
We compare our proposed MADUN with eight recent representative CS reconstruction methods, including two DNUNs and six DUNs.
The average PSNR reconstruction performances on Set11, CBSD68 and Urban100 data-set with respect to five CS ratios are summarized in Table~\ref{tab:fixed_matrix}. The models of ReconNet \cite{Kulkarni2016ReconNetNR}, ISTA-Net$^+$ \cite{zhang2018ista}, DPA-Net \cite{sun2020dual}, IRCNN \cite{zhang2017learning},  MAC-Net \cite{chenlearning} and iPiano-Net \cite{su2020ipiano} are trained by the same methods and training datasets with the corresponding works, and DPDNN \cite{dong2018denoising} and GDN \cite{gilton2019neumann} utilize the same training dataset with our method due to no CS reconstruction task in their original works. One can observe that our MADUN outperforms all the other competing methods in PSNR and SSIM across all the cases. 
Figure~\ref{fig:fixed_matrix} further show the visual comparisons on challenging images on Set11 dataset and Urban100 dataset, respectively. Our MADUN generates images that are visually pleasant and faithful to the groundtruth. 

\subsubsection{\textbf{Jointly Learned Matrix}}
We mimic the CS sampling process $\mathbf{y} = \mathbf{\Phi}\mathbf{x}\in\mathbb{R}^M$ using a convolution layer, and utilize another convolution layer with $N$ filters from $\mathbf{\Phi^{\top}}\in\mathbb{R}^{N \times M}$ to obtain initialization $\mathbf{\Phi^\top}\mathbf{y}$, like \cite{zhang2020optimization}. We compare MADUN with some methods which jointly learn the sampling matrix and the reconstruction process. 
Table~\ref{tab:joint_matrix} presents quantitative results on Set11 and CBSD68 dataset. As we can see, our MADUN achieves the best performance on all ratios. Figure~\ref{fig:joint_matrix} further shows the visual comparisons on challenging images, one can see that MADUN recovers richer image structures and texture details than all the other methods. 

\subsection{Application to Compressive Sensing MRI}
To demonstrate the generality of MADUN, we directly extend MA-DUN to the speciﬁc problem of CS-MRI reconstruction, which aims at reconstructing MR images from a small number of under-sampled data in k-space. In this application, we set the sampling process $\mathbf{\Phi}$ in Eq.~\eqref{eq:ista-r} to $\mathbf{\Phi}=\mathbf{B}\mathbf{F}$, where $\mathbf{B}$ is an under-sampling matrix and $\mathbf{F}$ is the discrete Fourier transform. In this case, we compare against seven recent methods for the CS-MRI domain.
Utilizing the same training and testing brain medical images as ADMM-Net \cite{yang2018admm}, the CS-MRI results of MADUNs are summarized in Table~\ref{tab:brain600} for different CS ratios. From Table ~\ref{tab:brain600}, we can see that our proposed method outperforms the state-of-the-art methods on testing brain dataset with all tested CS ratios, especially when ratios are smaller. 

\section{Conclusion}
We propose a novel \textbf{M}emory-\textbf{A}ugmented \textbf{D}eep \textbf{U}nfolding \textbf{N}etwork (\textbf{MADUN}) for CS, which addresses the problem of the information-lossy short-term transmission between each two adjacent stages in traditional DUNs by establishing and strengthening both the short-term and long-term memory flows with large capacity. Our two types of memory augmentation mechanisms, dubbed High-throughput Short-term Memory (HSM) and Cross-stage Long-term Memory (CLM), are developed and integrated into our Memory-Augmented Proximal Mapping Module (MAPMM). HSM and CLM collaboratively achieve persistent and adaptive memory by bridging adjacent stages with the multi-channel signal path and developing the dependency of deep information across all cascading stages, respectively. Extensive CS experiments on both natural and MR images exhibit that the proposed MADUN achieves better signal balance based on both memory mechanisms and outperforms existing state-of-the-art deep network-based methods with large margins. In the future, we will further extend our MADUN to other image inverse problems and video applications.


\bibliographystyle{ACM-Reference-Format}
\bibliography{sample-base}


\end{document}